\documentclass[11pt,a4paper]{article}
\usepackage[hyperref]{acl2019}
\usepackage{times}
\usepackage{latexsym}
\usepackage{amsmath,amsfonts,amssymb,amsthm, bm}
\usepackage{multirow}
\usepackage{url}
\usepackage{array}
\usepackage{graphicx}
\usepackage{subcaption}
\usepackage{amssymb}
\usepackage{bbding}
\usepackage{color}
\usepackage{arydshln}
\usepackage{enumerate}
\usepackage{verbatim}

\usepackage{chngcntr}
\usepackage{pifont}
\aclfinalcopy 
\def\aclpaperid{889} 

\title{Multimodal Transformer for Unaligned Multimodal Language Sequences}

\author{Yao-Hung Hubert Tsai$^*$
\\Carnegie Mellon University \\ (*equal contribution) \And Shaojie Bai$^*$
\\Carnegie Mellon University \\ (*equal contribution) \And Paul Pu Liang
\\Carnegie Mellon University \AND J. Zico Kolter
\\Carnegie Mellon University \\ and Bosch Center for AI \And Louis-Philippe Morency
\\Carnegie Mellon University \And Ruslan Salakhutdinov
\\Carnegie Mellon University
\AND
\url{https://github.com/yaohungt/Multimodal-Transformer}
}

\date{}

\begin{document}
\maketitle
\begin{abstract}
Human language is often multimodal, which comprehends a mixture of natural language, facial gestures, and acoustic behaviors. However, two major challenges in modeling such multimodal human language time-series data exist: 1) inherent data non-alignment due to variable sampling rates for the sequences from each modality; and 2) long-range dependencies between elements across modalities. In this paper, we introduce the Multimodal Transformer (MulT) to generically address the above issues in an end-to-end manner without explicitly aligning the data. At the heart of our model is the directional pairwise crossmodal attention, which attends to interactions between multimodal sequences across distinct time steps and latently adapt streams from one modality to another. Comprehensive experiments on both aligned and non-aligned multimodal time-series show that our model outperforms state-of-the-art methods by a large margin. In addition, empirical analysis suggests that correlated crossmodal signals are able to be captured by the proposed crossmodal attention mechanism in MulT.
\end{abstract}

\section{Introduction}

Human language possesses not only spoken words but also nonverbal behaviors from vision (facial attributes) and acoustic (tone of voice) modalities~\cite{gibson1994tools}. This rich information provides us the benefit of understanding human behaviors and intents~\citep{Manning14thestanford}. Nevertheless, the heterogeneities across modalities often increase the difficulty of analyzing human language. For example, the receptors for audio and vision streams may vary with variable receiving frequency, and hence we may not obtain optimal mapping between them. A frowning face may relate to a pessimistically word spoken in the past. That is to say, multimodal language sequences often exhibit ``unaligned'' nature and require inferring long term dependencies across modalities, which raises a question on performing efficient multimodal fusion.

\begin{figure}[t!]
    \centering
    \includegraphics[width=0.5\textwidth]{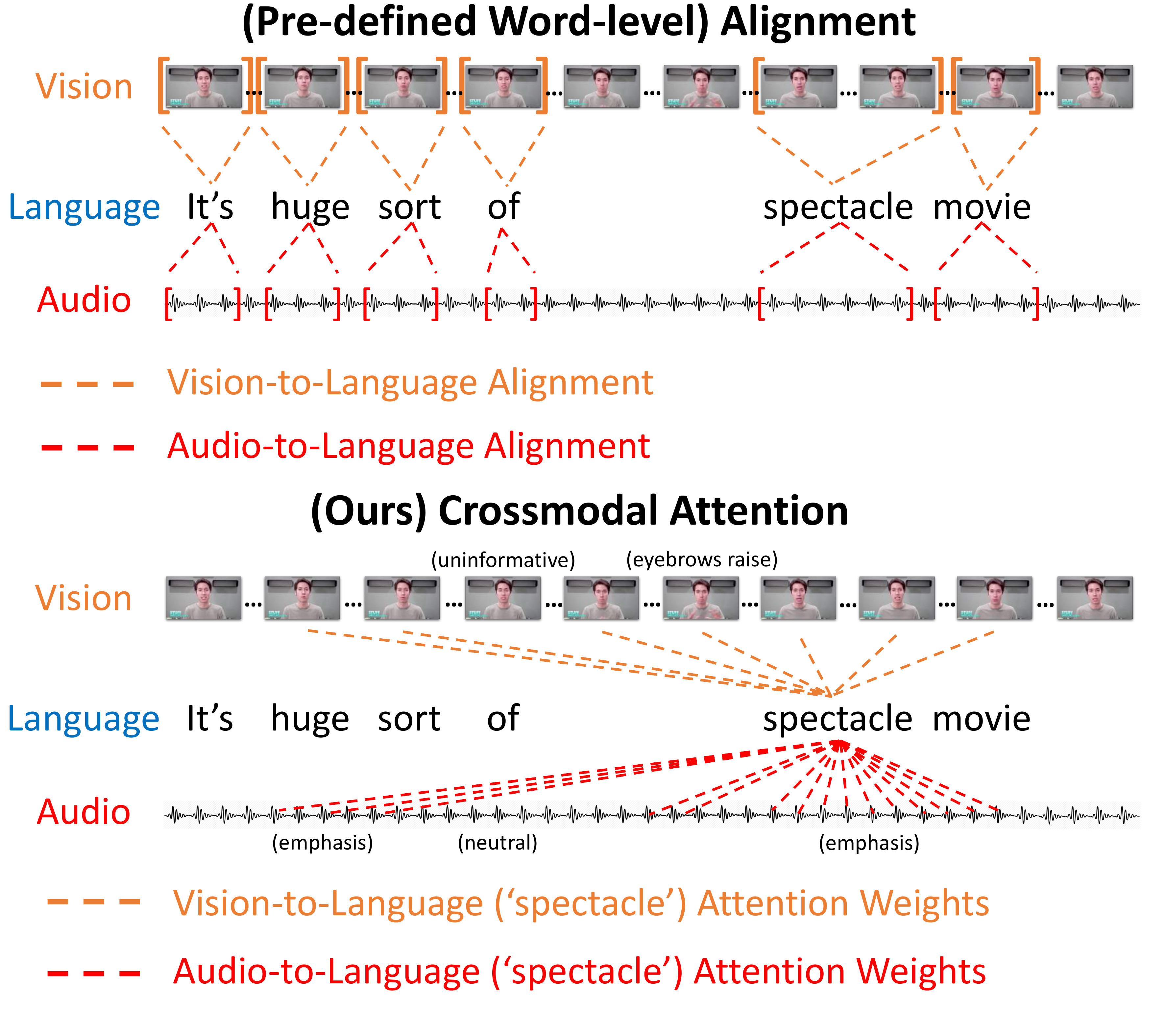}
    \vspace{-6mm}
    \caption{\small Example video clip from movie reviews. [Top]: Illustration of word-level alignment where video and audio features are averaged across the time interval of each spoken word. [Bottom] Illustration of crossmodal attention weights between text (``spectacle'') and vision/audio.}
    \label{fig:fig1}
    \vspace{-3mm}
\end{figure}

To address the above issues, in this paper we propose the Multimodal Transformer (MulT), an end-to-end model that extends the standard Transformer network~\cite{vaswani2017attention} to learn representations directly from unaligned multimodal streams. At the heart of our model is the crossmodal attention module, which attends to the crossmodal interactions at the scale of the entire utterances. This module latently adapts streams from one modality to another (e.g., $\mathrm{vision} \rightarrow \mathrm{language}$) by repeated reinforcing one modality's features with those from the other modalities, regardless of the need for alignment. In comparison, one common way of tackling unaligned multimodal sequence is by forced word-aligning before training~\cite{poria2017context, zadeh2018memory, zadeh2018multimodal, tsai2018learning,pham2018found,P18-1207}: manually preprocess the visual and acoustic features by aligning them to the resolution of words. These approaches would then model the multimodal interactions on the (already) aligned time steps and thus do not directly consider long-range crossmodal contingencies of the original features. We note that such word-alignment not only requires feature engineering that involves domain knowledge; but in practice, it may also not always be feasible, as it entails extra meta-information about the datasets (e.g., the exact time ranges of words or speech utterances). 
We illustrate the difference between the word-alignment and the crossmodal attention inferred by our model in Figure~\ref{fig:fig1}.

For evaluation, we perform a comprehensive set of experiments on three human multimodal language benchmarks: CMU-MOSI~\cite{zadeh2016multimodal}, CMU-MOSEI~\cite{zadeh2018multimodal}, and IEMOCAP~\cite{busso2008iemocap}. Our experiments show that MulT achieves the state-of-the-art (SOTA) results in not only the commonly evaluated word-aligned setting but also the more challenging unaligned scenario, outperforming prior approaches by a margin of 5\%-15\% on most of the metrics. In addition, empirical qualitative analysis further suggests that the crossmodal attention used by MulT is capable of capturing correlated signals across asynchronous modalities.

\section{Related Works}
\label{sec:rela}

\paragraph{Human Multimodal Language Analysis.}
Prior work for analyzing human multimodal language lies in the domain of inferring representations from multimodal sequences spanning language, vision, and acoustic modalities. Unlike learning multimodal representations from static domains such as image and textual attributes~\cite{ngiam2011multimodal,srivastava2012multimodal}, human language contains time-series and thus requires fusing time-varying signals~\cite{liang2018multimodal,tsai2018learning}. Earlier work used early fusion approach to concatenate input features from different modalities~\cite{lazaridou2015combining,ngiam2011multimodal} and showed improved performance as compared to learning from a single modality. More recently, more advanced models were proposed to learn representations of human multimodal language. For example,~\citet{P18-1207} used hierarchical attention strategies to learn multimodal representations, ~\citet{wang2018words} adjusted the word representations using accompanying non-verbal behaviors, ~\citet{pham2018found} learned robust multimodal representations using a cyclic translation objective, and~\citet{nips2018fusion} explored cross-modal autoencoders for audio-visual alignment. These previous approaches relied on the assumption that multimodal language sequences are already aligned in the resolution of words and considered only short-term multimodal interactions. In contrast, our proposed method requires no alignment assumption and defines crossmodal interactions at the scale of the entire sequences.

\paragraph{Transformer Network.}
\label{subsec:transformer}

Transformer network~\cite{vaswani2017attention} was first introduced for neural machine translation (NMT) tasks, where the encoder and decoder side each leverages a \emph{self-attention}~\cite{parikh2016decomposable, lin2017structured, vaswani2017attention} transformer. After each layer of the self-attention, the encoder and decoder are connected by an additional decoder sublayer where the decoder attends to each element of the source text for each element of the target text. We refer the reader to~\cite{vaswani2017attention} for a more detailed explanation of the model. In addition to NMT, transformer networks have also been successfully applied to other tasks, including language modeling~\citep{dai2018transformer,baevski2018adaptive}, semantic role labeling~\citep{D18-1548}, word sense disambiguation~\citep{tang2018self}, learning sentence representations~\citep{devlin2018bert}, and video activity recognition~\citep{wang2018non}.

This paper absorbs a strong inspiration from the NMT transformer to extend to a multimodal setting. Whereas the NMT transformer focuses on unidirectional \emph{translation} from source to target texts, human multimodal language time-series are neither as well-represented nor discrete as word embeddings, with sequences of each modality having vastly different frequencies. Therefore, we propose not to explicitly translate from one modality to the others (which could be extremely challenging), but to \emph{latently} adapt elements across modalities via the attention. Our model (MulT) therefore has no encoder-decoder structure, but it is built up from multiple stacks of pairwise and bidirectional crossmodal attention blocks that directly attend to low-level features (while removing the self-attention). Empirically, we show that our proposed approach improves beyond standard transformer on various human multimodal language tasks.

\section{Proposed Method}
\label{sec:method}

\begin{figure}[t!]
    \centering
    \includegraphics[width=0.5\textwidth]{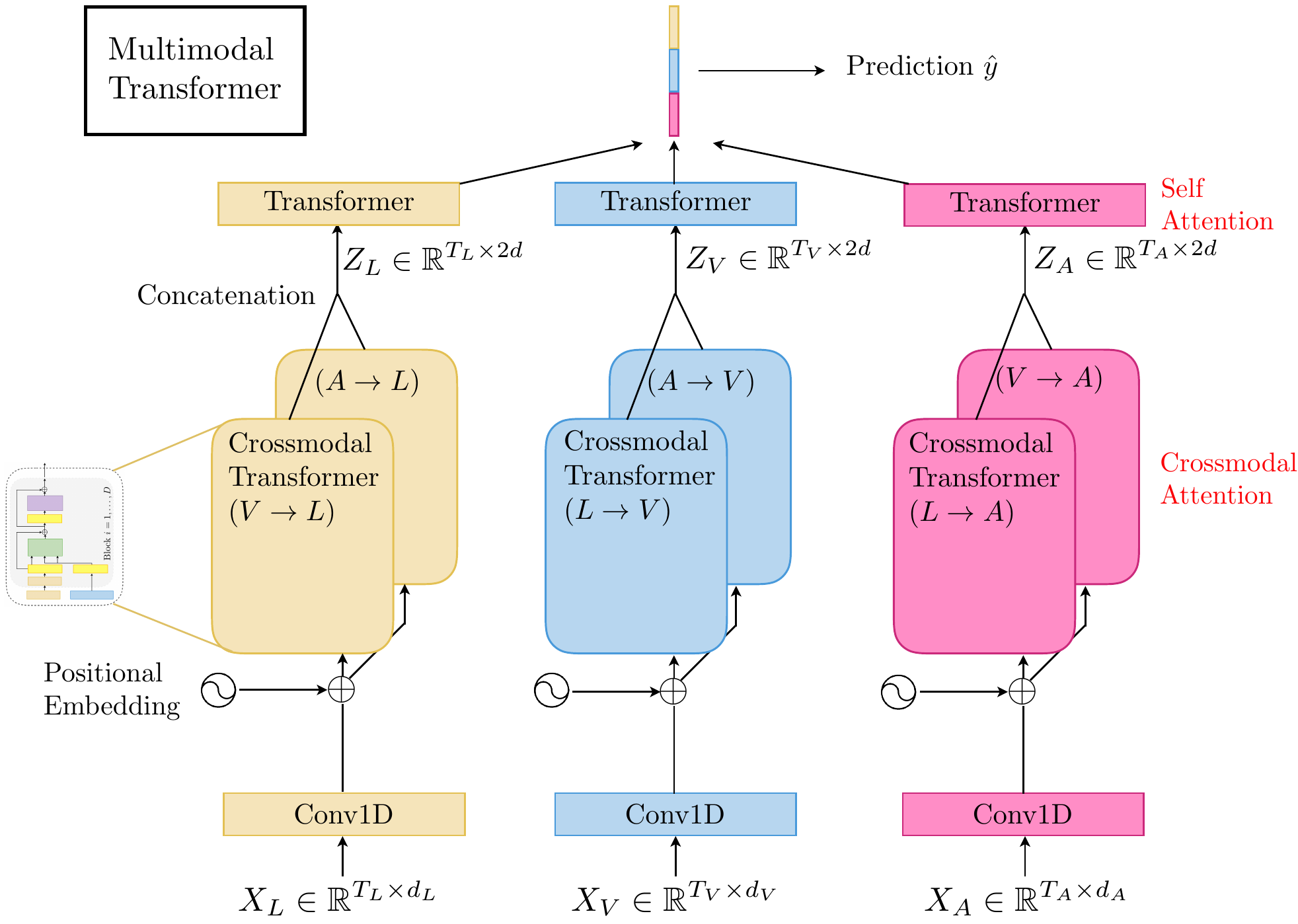}
    \vspace{-3mm}
    \caption{\small Overall architecture for MulT on modalities ($L, V, A$). The crossmodal transformers, which suggests latent crossmodal adaptations, are the core components of MulT for multimodal fusion.}
    \label{fig:overall}
\vspace{-3mm}
\end{figure}

In this section, we describe our proposed Multimodal Transformer (MulT) (Figure \ref{fig:overall}) for modeling unaligned multimodal language sequences. At the high level, MulT merges multimodal time-series via a feed-forward fusion process from multiple directional pairwise crossmodal transformers. Specifically, each crossmodal transformer (introduced in Section \ref{subsec:overall}) serves to repeatedly reinforce a \emph{target modality} with the low-level features from another \emph{source modality} by learning the attention across the two modalities' features. A MulT architecture hence models all pairs of modalities with such crossmodal transformers, followed by sequence models (e.g., self-attention transformer) that predicts using the fused features.

The core of our proposed model is crossmodal attention module, which we first introduce in Section~\ref{subsec:cm-attn}.
Then, in Section~\ref{subsec:overall} and~\ref{subsec:advantage}, we present in details the various ingredients of the MulT architecture (see Figure~\ref{fig:overall}) and discuss the difference between crossmodal attention and classical multimodal alignment.

\begin{figure*}[t]
    \vspace{-.1in}
    \centering
    \begin{subfigure}[b]{0.6\textwidth}
        \centering
        \includegraphics[width=0.85\textwidth]{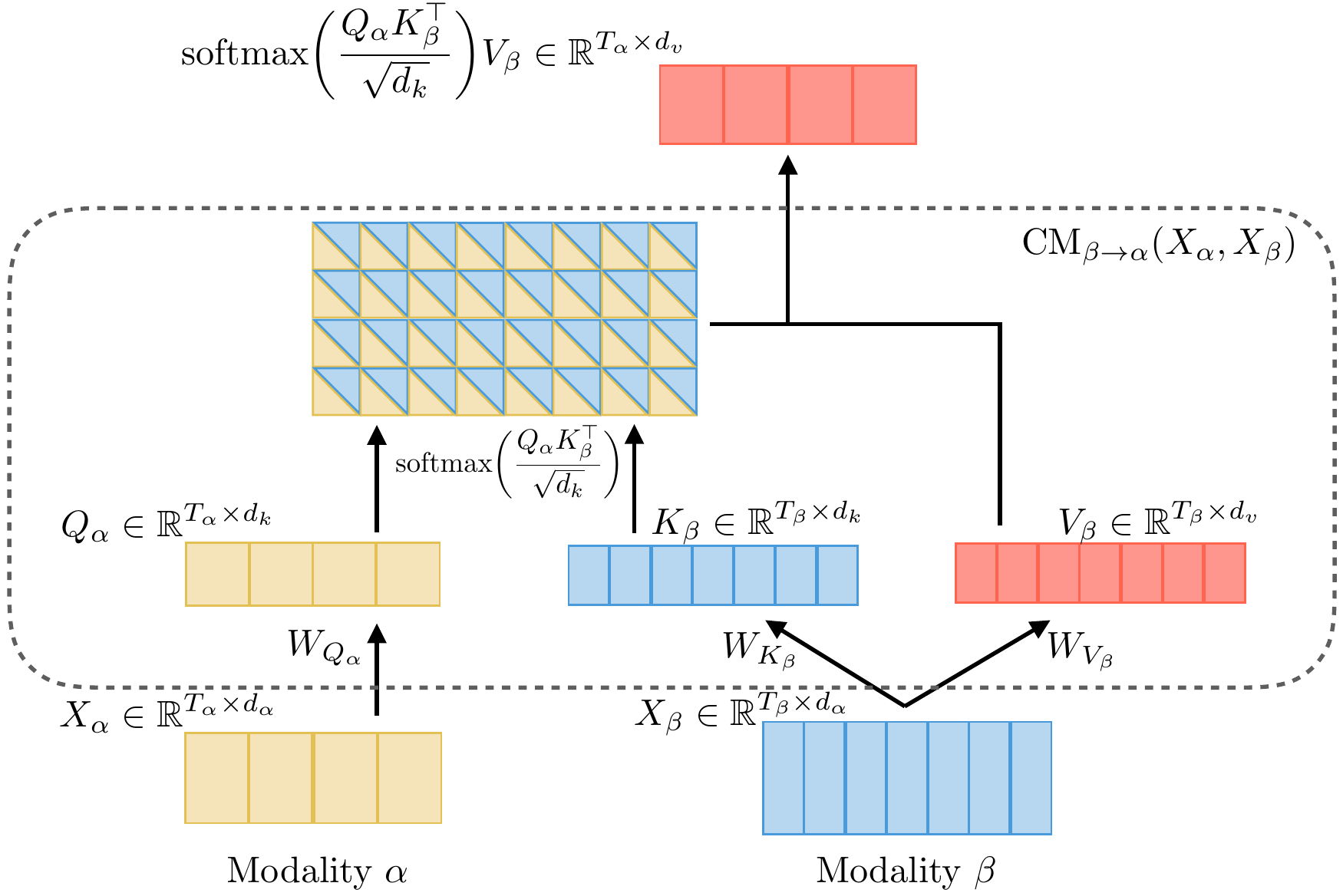} 
        \caption{\small Crossmodal attention $\text{CM}_{\beta \rightarrow \alpha}(X_\alpha, X_\beta)$ between sequences $X_\alpha$, $X_\beta$ from distinct modalities.}
        \label{fig:components-cm}
    \end{subfigure}
    ~
    \begin{subfigure}[b]{0.36\textwidth}
        \centering
        \includegraphics[width=\textwidth]{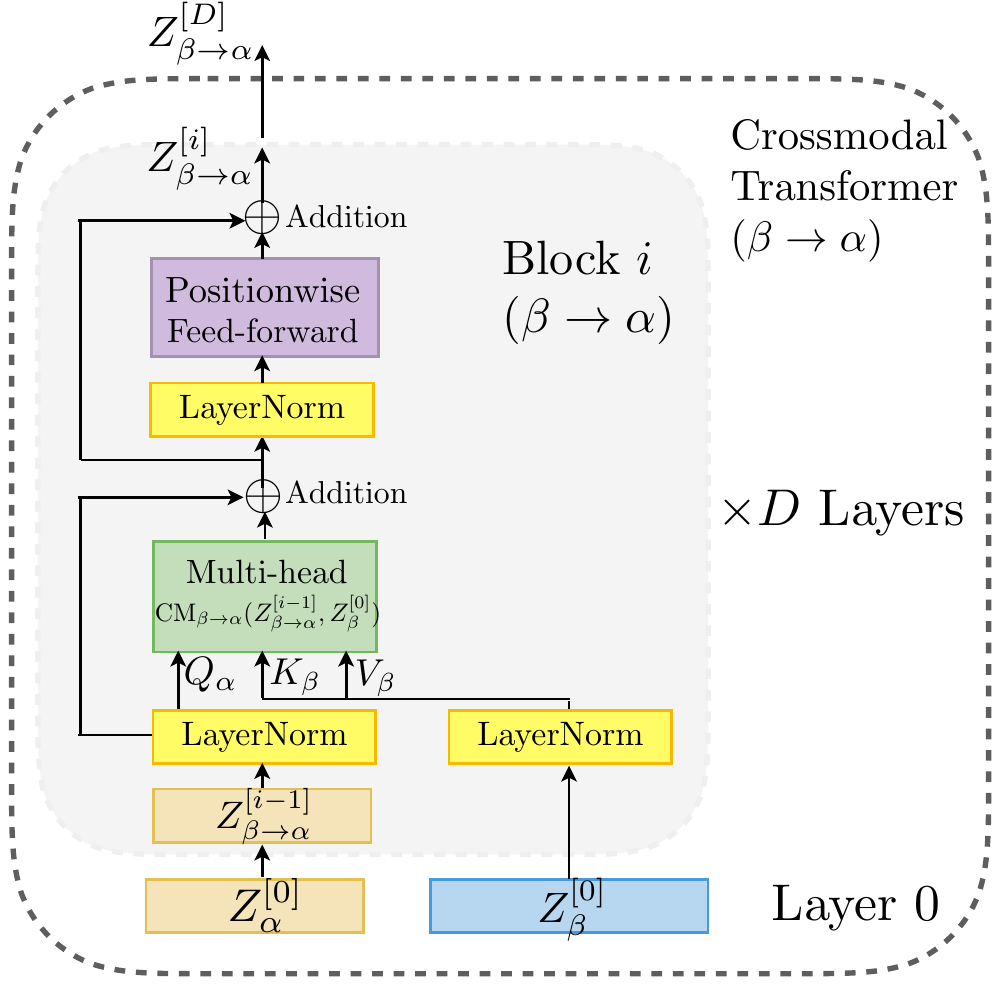}
        \caption{\small A crossmodal transformer is a deep stacking of several crossmodal attention blocks.}
        \label{fig:components-ct}
    \end{subfigure}
    \vspace{-.1in}
    \caption{\small Architectural elements of a crossmodal transformer between two time-series from modality $\alpha$ and $\beta$.}
    \label{fig:components}
\vspace{-3mm}
\end{figure*}



\subsection{Crossmodal Attention}
\label{subsec:cm-attn}

We consider two modalities $\alpha$ and $\beta$, with two (potentially non-aligned) sequences from each of them denoted $X_\alpha \in \mathbb{R}^{T_\alpha \times d_\alpha}$ and $X_\beta \in \mathbb{R}^{T_\beta \times d_\beta}$, respectively. For the rest of the paper, $T_{(\cdot)}$ and $d_{(\cdot)}$ are used to represent sequence length and feature dimension, respectively. Inspired by the decoder transformer in NMT~\citep{vaswani2017attention} that translates one language to another, we hypothesize a good way to fuse crossmodal information is providing a latent adaptation across modalities; i.e., $\beta$ to $\alpha$. Note that the modalities consider in our paper may span very different domains such as facial attributes and spoken words.

We define the $\mathrm{Query}$s as $Q_\alpha = X_\alpha W_{Q_\alpha}$, $\mathrm{Key}$s as $K_\beta = X_\beta W_{K_\beta}$, and $\mathrm{Value}$s as $V_\beta = X_\beta W_{V_\beta}$, where $W_{Q_\alpha} \in \mathbb{R}^{d_\alpha \times d_k}, W_{K_\beta} \in \mathbb{R}^{d_\beta \times d_k}$ and $W_{V_\beta} \in \mathbb{R}^{d_\beta \times d_v}$ are weights. The latent adaptation from $\beta$ to $\alpha$ is presented as the crossmodal attention $Y_\alpha := \text{CM}_{\beta \rightarrow \alpha}(X_\alpha, X_B) \in \mathbb{R}^{T_\alpha \times d_v}$:
\vspace{-2mm}
\begin{equation}
\label{eq:crossmodalBA}
\resizebox{0.7\hsize}{!}{
$
\begin{split}
    Y_\alpha &= \text{CM}_{\beta \rightarrow \alpha}(X_\alpha, X_\beta) \\
    &= \text{softmax}\left(\frac{Q_\alpha K_\beta^\top}{\sqrt{d_k}}\right) V_\beta
    \\
    &= \text{softmax}\left(\frac{X_\alpha W_{Q_\alpha} W_{K_\beta}^\top X_\beta^\top}{\sqrt{d_k}}\right) X_\beta W_{V_\beta}.
\end{split}
$}
\vspace{-3mm}
\end{equation}
Note that $Y_\alpha$ has the same length as $Q_\alpha$ (i.e., $T_\alpha$), but is meanwhile represented in the feature space of $V_\beta$. Specifically, the scaled (by $\sqrt{d_k}$) softmax in Equation~(\ref{eq:crossmodalBA}) computes a score matrix $\mathrm{softmax\,}(
\cdot) \in \mathbb{R}^{T_\alpha \times T_\beta}$, whose $(i,j)$-th entry measures the attention given by the $i$-th time step of modality $\alpha$ to the $j$-th time step of modality $\beta$. Hence, the $i$-th time step of $Y_\alpha$ is a weighted summary of $V_\beta$, with the weight determined by $i$-th row in $\mathrm{softmax}(\cdot)$. We call Equation~\eqref{eq:crossmodalBA} a \emph{single-head} crossmodal attention, which is illustrated in Figure \ref{fig:components-cm}. 

Following prior works on transformers~\citep{vaswani2017attention,chen2018thebest,devlin2018bert,dai2018transformer}, we add a residual connection to the crossmodal attention computation. Then, another positionwise feed-forward sublayer is injected to complete a \emph{crossmodal attention
block} (see Figure \ref{fig:components-ct}). Each crossmodal attention block adapts directly from the low-level feature sequence (i.e., $Z_\beta^{[0]}$ in Figure \ref{fig:components-ct}) and does not rely on self-attention, which makes it different from the NMT encoder-decoder architecture~\citep{vaswani2017attention, shaw2018self} (i.e., taking intermediate-level features). 
We argue that performing adaptation from low-level feature benefits our model to preserve the low-level information for each modality. 
We leave the empirical study for 
adapting from intermediate-level features (i.e., $Z_\beta^{[i-1]}$) in Ablation Study in Section~\ref{subsec:quant}.


\subsection{Overall Architecture}
\label{subsec:overall}

Three major modalities are typically involved in multimodal language sequences: language ($L$), video ($V$), and audio ($A$) modalities. We denote with ${X}_{\{L,V,A\}}\in \mathbb{R}^{T_{\{L,V,A\}} \times d_{\{L,V,A\}}}$ the input feature sequences (and the dimensions thereof) from these 3 modalities. With these notations, in this subsection, we describe in greater details the components of Multimodal Transformer and how crossmodal attention modules are applied.





\paragraph{Temporal Convolutions.} To ensure that each element of the input sequences has sufficient awareness of its neighborhood elements, we pass the input sequences through a 1D temporal convolutional layer:
\begin{equation}
\resizebox{\hsize}{!}{
$
\hat{X}_{\{L,V,A\}} = \text{Conv1D}(X_{\{L,V,A\}}, k_{\{L,V,A\}}) \in \mathbb{R}^{T_{\{L,V,A\}} \times d}
$
}
\end{equation}
where $k_{\{L,V,A\}}$ are the sizes of the convolutional kernels for modalities $\{L,V,A\}$, and $d$ is a common dimension. The convolved sequences are expected to contain the local structure of the sequence, which is important since the sequences are collected at different sampling rates. Moreover, since the temporal convolutions project the features of different modalities to the same dimension $d$, the dot-products are admittable in the crossmodal attention module.

\paragraph{Positional Embedding.} To enable the sequences to carry temporal information, following~\cite{vaswani2017attention}, we augment positional embedding (PE) to $\hat{X}_{\{L,V,A\}}$:
\vspace{-2mm}
\begin{equation}
    Z_{\{L,V,A\}}^{[0]} = \hat{X}_{\{L,V,A\}} + \text{PE}(T_{\{L,V,A\}}, d)
\end{equation}
where PE$(T_{\{L,V,A\}}, d) \in \mathbb{R}^{T_{\{L,V,A\}} \times d}$ computes the (fixed) embeddings for each position index, and $Z_{\{L,V,A\}}^{[0]}$ are the resulting low-level position-aware features for different modalities. We leave more details of the positional embedding to Appendix \ref{app:pe}.

\paragraph{Crossmodal Transformers.} Based on the crossmodal attention blocks, we design the crossmodal transformer that enables one modality for receiving information from another modality. In the following, we use the example for passing vision ($V$) information to language ($L$), which is denoted by ``$V\rightarrow L$''. We fix all the dimensions ($d_{\{\alpha,\beta,k,v \}}$) for each crossmodal attention block as $d$.

Each crossmodal transformer consists of $D$ layers of crossmodal attention blocks (see Figure \ref{fig:components-ct}). Formally, a crossmodal transformer computes feed-forwardly for $i=1, \dots, D$ layers:
\begin{equation}
\resizebox{\hsize}{!}{
$
\begin{split}
    Z_{V \rightarrow L}^{[0]} &= Z_L^{[0]} \\
    \hat{Z}_{V \rightarrow L}^{[i]} &= \text{CM}_{V \rightarrow L}^{[i], \text{mul}}(\text{LN}(Z_{V \rightarrow L}^{[i-1]}), \text{LN}(Z_V^{[0]})) + \text{LN}(Z_{V \rightarrow L}^{[i-1]}) \\
    Z_{V \rightarrow L}^{[i]} &= f_{\theta_{V \rightarrow L}^{[i]}}(\text{LN}(\hat{Z}_{V \rightarrow L}^{[i]})) + \text{LN}(\hat{Z}_{V \rightarrow L}^{[i]})
\end{split}
$}
\end{equation}
where $f_\theta$ is a positionwise feed-forward sublayer parametrized by $\theta$, and $\text{CM}_{V \rightarrow L}^{[i], \text{mul}}$ means a multi-head (see~\cite{vaswani2017attention} for more details) version of $\text{CM}_{V \rightarrow L}$ at layer $i$ (note: $d$ should be divisible by the number of heads). LN means layer normalization~\citep{ba2016layer}.

In this process, each modality keeps updating its sequence via low-level external information from the multi-head crossmodal attention module. At every level of the crossmodal attention block, the low-level signals from source modality are transformed to a different set of $\mathrm{Key}$/$\mathrm{Value}$ pairs to interact with the target modality. Empirically, we find that the crossmodal transformer learns to correlate meaningful elements across modalities (see Section \ref{sec:exp} for details). The eventual MulT is based on modeling every pair of crossmodal interactions. Therefore, with $3$ modalities (i.e., $L,V,A$) in consideration, we have $6$ crossmodal transformers in total (see Figure \ref{fig:overall}).

\paragraph{Self-Attention Transformers and Prediction.} As a final step, we concatenate the outputs from the crossmodal transformers that share the same target modality to yield $Z_{\{L,V,A\}} \in \mathbb{R}^{T_{\{L,V,A\}} \times 2d}$. For example, $Z_L = [Z_{V\rightarrow L}^{[D]}; Z_{A\rightarrow L}^{[D]}]$. Each of them is then passed through a sequence model to collect temporal information to make predictions. We choose the self-attention transformer~\citep{vaswani2017attention}. Eventually, the last elements of the sequences models are extracted to pass through fully-connected layers to make predictions. 

\begin{figure}[t!]
    \centering
    \vspace{-5mm}
    \includegraphics[width=0.49\textwidth]{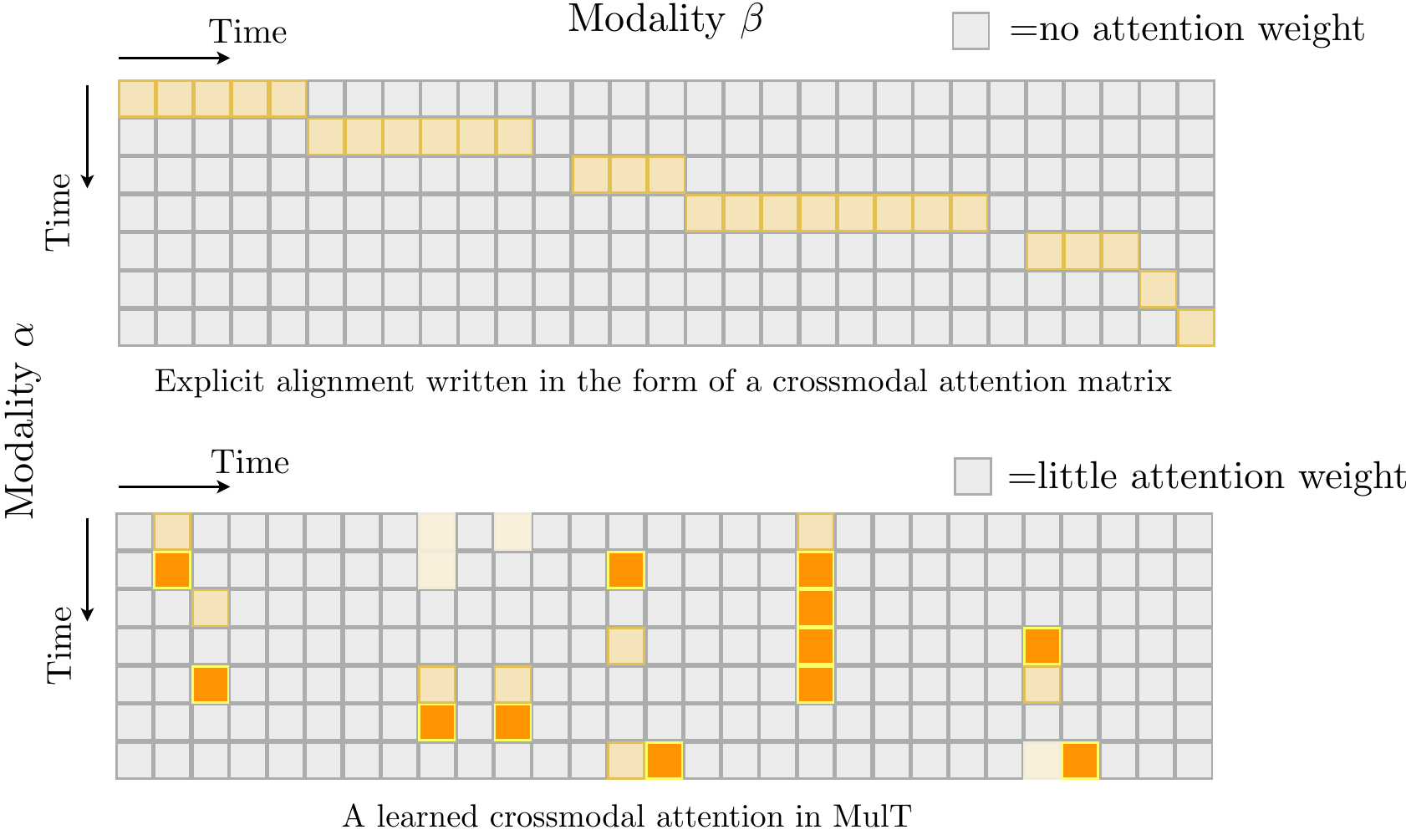}
    \vspace{-7mm}
    \caption{\small An example of visualizing alignment using attention matrix from modality $\beta$ to $\alpha$. Multimodal alignment is a special (monotonic) case for crossmodal attention.
    }
    \label{fig:align-attn}
    \vspace{-3mm}
\end{figure}

\subsection{Discussion about Attention \& Alignment}
\label{subsec:advantage}

When modeling unaligned multimodal language sequences, MulT relies on crossmodal attention blocks to merge signals across modalities. While the multimodal sequences were (manually) aligned to the same length in prior works before training~\citep{zadeh2018multimodal, liang2018multimodal, tsai2018learning, pham2018found, wang2018words}, we note that MulT looks at the non-alignment issue through a completely different lens. Specifically, for MulT, the correlations between elements of multiple modalities are purely based on attention. In other words, MulT does not handle modality non-alignment by (simply) aligning them; instead, the crossmodal attention encourages the model to directly attend to elements in other modalities where strong signals or relevant information is present. As a result, MulT can capture long-range crossmodal contingencies in a way that conventional alignment could not easily reveal. Classical crossmodal alignment, on the other hand, can be expressed as a special (step diagonal) crossmodal attention matrix  (i.e., monotonic attention~\cite{yu2016online}). We illustrate their differences in Figure~\ref{fig:align-attn}.

\section{Experiments}
\label{sec:exp}

In this section, we empirically evaluate the Multimodal Transformer (MulT) on three datasets that are frequently used to benchmark human multimodal affection recognition in prior works~\citep{pham2018found,tsai2018learning,liang2018multimodal}. Our goal is to compare MulT with prior competitive approaches on both \emph{word-aligned} (by word, which almost all prior works employ) and \emph{unaligned} (which is more challenging, and which MulT is generically designed for) multimodal language sequences.

\newcolumntype{K}[1]{>{\centering\arraybackslash}p{#1}}
\begin{table}[t!]
\vspace{-5mm}
\caption{\small Results for multimodal sentiment analysis on CMU-MOSI with aligned and non-aligned multimodal sequences. ${}^h$ means higher is better and ${}^\ell$ means lower is better. EF stands for early fusion, and LF stands for late fusion.}
\begin{center}
\fontsize{7}{10}\selectfont
\setlength\tabcolsep{0.7pt}
\vspace{-4mm}
\begin{tabular}{|c||*{5}{K{0.72cm}}|}
\hline
Metric         & Acc$_7^h$  & Acc$_2^h$  & F1$^h$    & MAE$^\ell$   & Corr$^h$  \\ \hline \hline
\multicolumn{6}{|c|}{\textcolor{blue}{(Word Aligned)} CMU-MOSI Sentiment}     \\ \hline \hline
EF-LSTM        & 33.7  & 75.3  & 75.2  & 1.023 & 0.608 \\
LF-LSTM        & 35.3     & 76.8  & 76.7  & 1.015 & 0.625 \\
RMFN~\cite{liang2018multimodal}     & 38.3  & 78.4  & 78.0  & 0.922 & 0.681 \\
MFM~\cite{tsai2018learning}         & 36.2  & 78.1  & 78.1  & 0.951 & 0.662 \\
RAVEN~\cite{wang2018words}          & 33.2     & 78.0  & 76.6    & 0.915 & \textbf{0.691} \\
MCTN~\cite{pham2018found}           & 35.6     & 79.3  & 79.1  & 0.909 & 0.676 \\ \hline \hline
MulT (ours)    & \bf{40.0}     & \bf{83.0}  & \bf{82.8}  & \bf{0.871} & \bf{0.698} \\ \hline \hline
\multicolumn{6}{|c|}{\textcolor{red}{(Unaligned)} CMU-MOSI Sentiment} \\ \hline \hline
CTC~\citep{graves2006connectionist} + EF-LSTM  & 31.0     & 73.6  & 74.5  & 1.078 & 0.542 \\
LF-LSTM        & 33.7     & 77.6  & 77.8  & 0.988 & 0.624 \\
CTC + MCTN~\cite{pham2018found}     & 32.7     & 75.9  & 76.4  & 0.991 & 0.613 \\
CTC + RAVEN~\cite{wang2018words}    & 31.7     & 72.7  & 73.1  & 1.076 & 0.544 \\ \hline \hline
MulT (ours)    & \bf{39.1}     & \bf{81.1}  & \bf{81.0}  & \bf{0.889} & \bf{0.686}  \\
\hline
\end{tabular}
\end{center}
\label{tbl:mosi}
\vspace{-4mm}
\end{table}
\newcolumntype{K}[1]{>{\centering\arraybackslash}p{#1}}
\begin{table}[t!]
\vspace{-.1mm}
\caption{\small Results for multimodal sentiment analysis on (relatively large scale) CMU-MOSEI with aligned and non-aligned multimodal sequences.}
\begin{center}
\fontsize{7}{10}\selectfont
\setlength\tabcolsep{0.7pt}
\vspace{-4mm}
\begin{tabular}{|c||*{5}{K{0.72cm}}|}
\hline
Metric         & Acc$_7^h$  & Acc$_2^h$  & F1$^h$    & MAE$^\ell$   & Corr$^h$  \\ \hline \hline
\multicolumn{6}{|c|}{\textcolor{blue}{(Word Aligned)} CMU-MOSEI Sentiment}     \\ \hline \hline
EF-LSTM        & 47.4  & 78.2  & 77.9  & 0.642 & 0.616 \\
LF-LSTM        & 48.8    & 80.6  & 80.6  & 0.619 & 0.659 \\
Graph-MFN~\cite{zadeh2018multimodal}           & 45.0  & 76.9  & 77.0  & 0.71 & 0.54 \\
RAVEN~\cite{wang2018words}          & 50.0     & 79.1  & 79.5     & 0.614 & 0.662 \\ 
MCTN~\cite{pham2018found}           & 49.6     & 79.8  & 80.6  & 0.609 & 0.670 \\ \hline \hline
MulT (ours)           & \bf{51.8}     & \bf{82.5}  & \bf{82.3}  & \bf{0.580} & \bf{0.703} \\ \hline \hline
\multicolumn{6}{|c|}{\textcolor{red}{(Unaligned)} CMU-MOSEI Sentiment} \\ \hline \hline
CTC~\citep{graves2006connectionist} + EF-LSTM  & 46.3  &  76.1  & 75.9  & 0.680 & 0.585   \\
LF-LSTM        & 48.8     & 77.5  & 78.2    &  0.624 & 0.656 \\
CTC + RAVEN~\cite{wang2018words}    & 45.5     & 75.4  & 75.7  & 0.664 & 0.599 \\ 
CTC + MCTN~\cite{pham2018found}     & 48.2     & 79.3  & 79.7  & 0.631 &  0.645 \\ \hline \hline
MulT (ours)           & \textbf{50.7}     & \textbf{81.6}  & \textbf{81.6}  & \textbf{0.591} & \textbf{0.694} \\\hline
\end{tabular}
\end{center}
\label{tbl:mosei}
\vspace{-4mm}
\end{table}

\newcolumntype{K}[1]{>{\centering\arraybackslash}p{#1}}
\begin{table*}[t!]
\vspace{-5mm}
\caption{\small Results for multimodal emotions analysis on IEMOCAP with aligned and non-aligned multimodal sequences.}
\begin{center}
\fontsize{7}{10}\selectfont
\setlength\tabcolsep{0.7pt}
\vspace{-4mm}
\begin{tabular}{|c||*{8}{K{0.83cm}}|}
\hline
Task          & \multicolumn{2}{c}{Happy} & \multicolumn{2}{c}{Sad} & \multicolumn{2}{c}{Angry} & \multicolumn{2}{c|}{Neutral} \\
Metric        & Acc$^h$         & F1$^h$          & Acc$^h$        & F1$^h$         & Acc$^h$         & F1$^h$          & Acc$^h$           & F1$^h$           \\ \hline \hline
\multicolumn{9}{|c|}{\textcolor{blue}{(Word Aligned)} IEMOCAP Emotions}                                                                               \\ \hline  \hline
EF-LSTM       & 86.0        & 84.2        & 80.2       & 80.5       & 85.2        & 84.5        & 67.8          & 67.1         \\
LF-LSTM       & 85.1        & 86.3        & 78.9       & 81.7       & 84.7        & 83.0        & 67.1          & 67.6         \\
RMFN~\cite{liang2018multimodal}          & 87.5        & 85.8        & 83.8       & 82.9       & 85.1        & 84.6        & 69.5          & 69.1         \\
MFM~\cite{tsai2018learning}           & 90.2        & 85.8        & \textbf{88.4}    & \textbf{86.1}       & \textbf{87.5}        & 86.7        & 72.1          & 68.1         \\
RAVEN~\cite{wang2018words}         & 87.3        & 85.8        & 83.4       & 83.1     &  \textbf{87.3}        & 86.7        & 69.7          & 69.3         \\
MCTN~\cite{pham2018found}          & 84.9          & 83.1           & 80.5        & 79.6          & 79.7          & 80.4           & 62.3            & 57.0            \\ \hline  \hline
MulT (ours)          & \bf{90.7}        & \bf{88.6}        & 86.7       & \textbf{86.0}       & \textbf{87.4}        & \textbf{87.0}        & \textbf{72.4}          & \bf{70.7}         \\ \hline  \hline
\multicolumn{9}{|c|}{\textcolor{red}{(Unaligned)} IEMOCAP Emotions}                                                                           \\ \hline  \hline
CTC~\cite{graves2006connectionist} + EF-LSTM  &  76.2  &  75.7  & 70.2  & 70.5  & 72.7  &  67.1  &  58.1 &  57.4 \\
LF-LSTM       & 72.5  &  71.8  &  72.9  &  70.4   &  68.6  &  67.9  &   59.6   &   56.2     \\
CTC + RAVEN~\cite{wang2018words}  &  77.0  &  76.8  & 67.6  &  65.6  & 65.0  & 64.1  &  \bf{62.0}  &  \bf{59.5} \\
CTC + MCTN~\cite{pham2018found}   &  80.5  & 77.5  & 72.0  & 71.7 &   64.9  &  65.6  &  49.4  &  49.3 \\  \hline \hline
MulT (ours)         & \bf{84.8}        & \bf{81.9}        & \bf{77.7}       & \bf{74.1}       & \bf{73.9}        & \bf{70.2}        & \bf{62.5}          & \bf{59.7}      \\  \hline
\end{tabular}
\end{center}
\label{tbl:iemocap}
\vspace{-4mm}
\end{table*}

\subsection{Datasets and Evaluation Metrics}

Each task consists of a \emph{word-aligned} (processed in the same way as in prior works) and an \emph{unaligned} version. For both versions, the multimodal features are extracted from the textual (GloVe word embeddings~\cite{pennington2014glove}), visual (Facet~\cite{emotient}), and acoustic (COVAREP~\cite{degottex2014covarep}) data modalities. A more detailed introduction to the features is included in Appendix~\ref{sec:feat}. 

For the word-aligned version, following~\cite{zadeh2018memory, tsai2018learning, pham2018found}, we first use P2FA~\cite{P2FA} to obtain the aligned timesteps (segmented w.r.t. words) for audio and vision streams, and we then perform averaging on the audio and vision features within these time ranges. All sequences in the word-aligned case have length 50. The process remains the same across all the datasets. On the other hand, for the unaligned version, we keep the original audio and visual features as extracted, without any word-segmented alignment or manual subsampling. As a result, the lengths of each modality vary significantly, where audio and vision sequences may contain up to $>1,000$ time steps. We elaborate on the three tasks below.


\paragraph{CMU-MOSI \& MOSEI.} CMU-MOSI~\cite{zadeh2016multimodal} is a human multimodal sentiment analysis dataset consisting of 2,199 short monologue video clips (each lasting the duration of a sentence). Acoustic and visual features of CMU-MOSI are extracted at a sampling rate of $12.5$ and $15$ Hz, respectively (while textual data are segmented per word and expressed as discrete word embeddings). Meanwhile, CMU-MOSEI~\cite{zadeh2018multimodal} is a sentiment and emotion analysis dataset made up of 23,454 movie review video clips taken from YouTube (about $10 \times$ the size of CMU-MOSI). The unaligned CMU-MOSEI sequences are extracted at a sampling rate of $20$ Hz for acoustic and $15$ Hz for vision signals. 

For both CMU-MOSI and CMU-MOSEI, each sample is labeled by human annotators with a sentiment score from -3 (strongly negative) to 3 (strongly positive). We evaluate the model performances using various metrics, in agreement with those employed in prior works: 7-class accuracy (i.e., Acc$_7$: sentiment score classification in $\mathbb{Z} \cap [-3,3]$), binary accuracy (i.e., Acc$_2$: positive/negative sentiments), F1 score, mean absolute error (MAE) of the score, and the correlation of the model's prediction with human. Both tasks are frequently used to benchmark models' ability to fuse multimodal (sentiment) information~\citep{poria2017context, zadeh2018memory, liang2018multimodal, tsai2018learning, pham2018found, wang2018words}.

\paragraph{IEMOCAP.} IEMOCAP~\cite{busso2008iemocap} consists of 10K videos for human emotion analysis. As suggested by~\citet{wang2018words}, 4 emotions (happy, sad, angry and neutral) were selected for emotion recognition. Unlike CMU-MOSI and CMU-MOSEI, this is a multilabel task (e.g., a person can be sad and angry simultaneously). Its multimodal streams consider fixed sampling rate on audio ($12.5$ Hz) and vision ($15$ Hz) signals. We follow~\cite{poria2017context, wang2018words, tsai2018learning} to report the binary classification accuracy
and the F1 score of the predictions.

\subsection{Baselines}

We choose Early Fusion LSTM (EF-LSTM) and Late Fusion LSTM (LF-LSTM) as baseline models, as well as Recurrent Attended Variation Embedding Network (RAVEN)~\cite{wang2018words} and Multimodal Cyclic Translation Network (MCTN)~\cite{pham2018found}, that achieved SOTA results on various word-aligned human multimodal language tasks. To compare the models comprehensively, we adapt the \textit{connectionist temporal classification} (CTC)~\cite{graves2006connectionist} method to the prior approaches (e.g., EF-LSTM, MCTN, RAVEN) that cannot be applied directly to the unaligned setting. Specifically, these models train to optimize the CTC alignment objective and the human multimodal objective simultaneously. We leave more detailed treatment of the CTC module to Appendix~\ref{app:ctc}. For fair comparisons, we control the number of parameters of all models to be approximately the same. The hyperparameters are reported in Appendix~\ref{app:hyper}. \footnote{All experiments are conducted on 1 GTX-1080Ti GPU. The code for our model and experiments can be found in \url{https://github.com/yaohungt/Multimodal-Transformer}}


\subsection{Quantitative Analysis}
\label{subsec:quant}

\paragraph{Word-Aligned Experiments.} 
We first evaluate MulT on the \emph{word-aligned sequences}--- the ``home turf'' of prior approaches modeling human multimodal language~\cite{sheikh2018sentiment,tsai2018learning,pham2018found,wang2018words}. The upper part of the Table~\ref{tbl:mosi},~\ref{tbl:mosei}, and~\ref{tbl:iemocap} show the results of MulT and baseline approaches on the word-aligned task. With similar model sizes (around 200K parameters), MulT outperforms the other competitive approaches on different metrics on all tasks, with the exception of the ``sad'' class results on IEMOCAP. 

\begin{figure}[t!]
    \vspace{-7mm}
    \centering
    \includegraphics[width=0.49\textwidth]{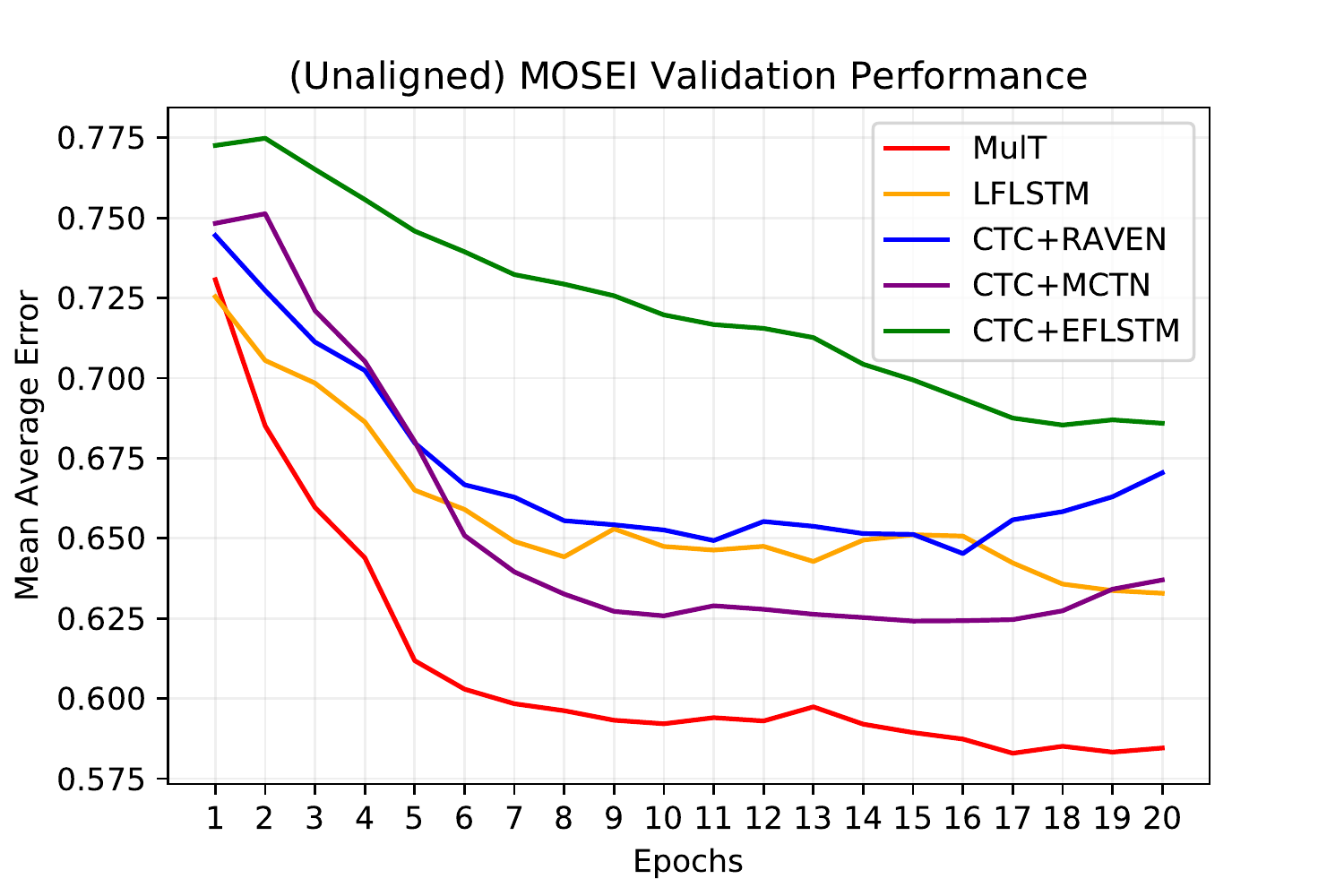}
    \vspace{-8mm}
    \caption{\small Validation set convergence of MulT when compared to other baselines on the \textcolor{red}{unaligned} CMU-MOSEI task.}
    \label{fig:cmu-mosei}
    \vspace{-4mm}
\end{figure}

\begin{figure*}[t!]
\vspace{-.1in}
\centering
\includegraphics[width=.86\textwidth]{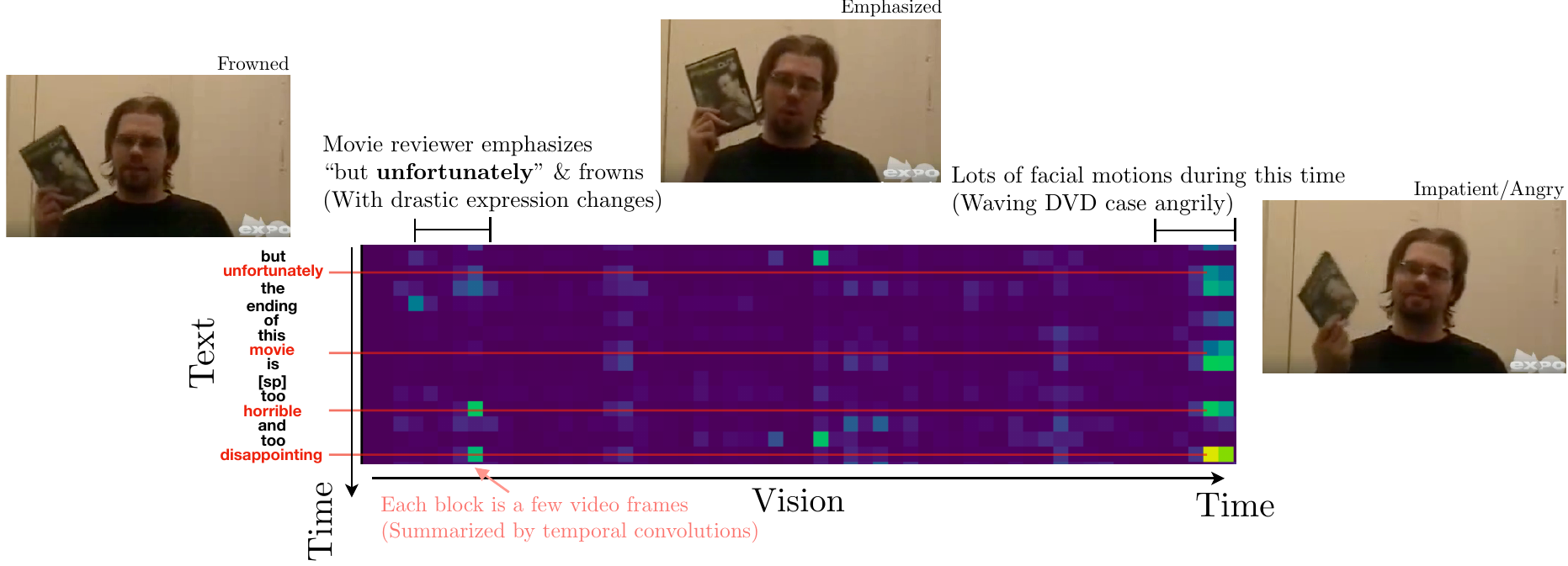} \\
\vspace{-3mm}
\caption{\small Visualization of sample crossmodal attention weights from layer 3 of $[V \rightarrow L]$ crossmodal transformer on CMU-MOSEI. We found that the crossmodal attention has learned to correlate certain meaningful words (e.g., ``movie'', ``disappointing'') with segments of stronger visual signals (typically stronger facial motions or expression change), despite the lack of alignment between original $L/V$ sequences. Note that due to temporal convolution, each textual/visual feature contains the representation of nearby elements.}
\label{fig:attn}
\vspace{-3mm}
\end{figure*}

\paragraph{Unaligned Experiments.} 
Next, we evaluate MulT on the same set of datasets in the unaligned setting. Note that MulT can be directly applied to unaligned multimodal stream, while the baseline models (except for LF-LSTM) require the need of additional alignment module (e.g., CTC module).

The results are shown in the bottom part of Table~\ref{tbl:mosi},~\ref{tbl:mosei}, and~\ref{tbl:iemocap}. On the three benchmark datasets, MulT improves upon the prior methods (some with CTC) by 10\%-15\% on most attributes. Empirically, we find that MulT converges faster to better results at training when compared to other competitive approaches (see Figure~\ref{fig:cmu-mosei}). In addition, while we note that in general there is a performance drop on all models when we shift from the word-aligned to unaligned multimodal time-series, the impact MulT takes is much smaller than the other approaches. We hypothesize such performance drop occurs because the asynchronous (and much longer) data streams introduce more difficulty in recognizing important features and computing the appropriate attention.

\paragraph{Ablation Study.} 

\newcolumntype{K}[1]{>{\centering\arraybackslash}p{#1}}

\begin{table}[t!]
\vspace{-5mm}
\caption{\small An ablation study on the benefit of MulT's crossmodal transformers using CMU-MOSEI.).
}
\label{tbl:abla}
\fontsize{7}{10}\selectfont
\vspace{-2mm}
\begin{tabular}{|c||*{5}{K{0.48cm}}|}
\hline
                                 & \multicolumn{5}{c|}{\textcolor{red}{(Unaligned)} CMU-MOSEI} \\
 Description                     & \multicolumn{5}{c|}{Sentiment}           \\
                                 & Acc$_7^h$   & Acc$_2^h$  & F1$^h$  & MAE$^\ell$  & Corr$^h$  \\ \hline \hline
\multicolumn{6}{|c|}{Unimodal Transformers}       \\ \hline \hline
Language only                   &  46.5  &   77.4  &   78.2  &   0.653  &  0.631  \\
Audio only                      &  41.4  &   65.6   &   68.8  &   0.764  &  0.310  \\
Vision only                     &  43.5  &   66.4   &   69.3  &   0.759  &  0.343  \\ \hline \hline
\multicolumn{6}{|c|}{Late Fusion by using Multiple Unimodal Transformers}   \\ \hline \hline
LF-Transformer                     &  47.9  &   78.6   &   78.5  &   0.636  &  0.658 
\\ \hline \hline
\multicolumn{6}{|c|}{Temporally Concatenated Early Fusion Transformer}   \\ \hline \hline
EF-Transformer                     &  47.8  &   78.9   &   78.8  &   0.648  &  0.647 
\\ \hline \hline
\multicolumn{6}{|c|}{Multimodal Transfomers} \\ \hline \hline
Only $[V,A \rightarrow L]$ (ours)     &  \textbf{50.5}   &   80.1   &   80.4   &  0.605  & 0.670         \\
Only $[L,A \rightarrow V]$ (ours)     &  48.2  &  79.7  &  80.2   &  0.611    &  0.651  \\
Only $[L,V \rightarrow A]$ (ours)     &  47.5  &  79.2  &  79.7   &  0.620    &  0.648         \\
\multirow{2}{*}{\shortstack[c]{MulT mixing intermediate-\\level features (ours)}}       &  \multirow{2}{*}{50.3}   &   \multirow{2}{*}{80.5}   &   \multirow{2}{*}{80.6}   &  \multirow{2}{*}{0.602}   &  \multirow{2}{*}{0.674}  \\
& & & & & \\
MulT (ours)                         &  \textbf{50.7}   &   \textbf{81.6}   &   \textbf{81.6}   &  \textbf{0.591}   &  \textbf{0.691}  \\ \hline
\end{tabular}
\vspace{-5mm}
\end{table}

To further study the influence of the individual components in MulT, we perform comprehensive ablation analysis using the unaligned version of CMU-MOSEI. The results are shown in Table~\ref{tbl:abla}. 

First, we consider the performance for only using unimodal transformers (i.e., language, audio or vision only). We find that the language transformer outperforms the other two by a large margin. For example, for the Acc$_2^h$ metric, the model improves from $65.6$ to $77.4$ when comparing audio only to language only unimodal transformer. This fact aligns with the observations in prior work~\cite{pham2018found}, where the authors found that a good language network could already achieve good performance at inference time. 

Second, we consider 1) a late-fusion transformer that feature-wise concatenates the last elements of three self-attention transformers; and 2) an early-fusion self-attention transformer that takes in a temporal concatenation of three asynchronous sequences $[\hat{X}_L, \hat{X}_V, \hat{X}_A] \in \mathbb{R}^{(T_L + T_V + T_A) \times d_q}$ (see Section~\ref{subsec:overall}). Empirically, we find that both EF- and LF-Transformer (which fuse multimodal signals) outperform unimodal transformers. 

Finally, we study the importance of individual crossmodal transformers according to the target modalities (i.e., using $[V, A \rightarrow L]$, $[L, A \rightarrow V]$, or $[L, V \rightarrow A]$ network). As shown in Table \ref{tbl:abla}, we find crossmodal attention modules consistently improve over the late- and early-fusion transformer models in most metrics on unaligned CMU-MOSEI. In particular, among the three crossmodal transformers, the one where language($L$) is the target modality works best. We also additionally study the effect of adapting intermediate-level instead of the low-level features from source modality in crossmodal attention blocks (similar to the NMT encoder-decoder architecture but without self-attention; see Section \ref{subsec:cm-attn}). While MulT leveraging intermediate-level features still outperform models in other ablative settings, we empirically find adapting from low-level features works best. The ablations suggest that crossmodal attention concretely benefits MulT with better representation learning.

\subsection{Qualitative Analysis}
\label{subsec:deeper}

To understand how crossmodal attention works while modeling unaligned multimodal data, we empirically inspect what kind of signals MulT picks up by visualizing the attention activations. Figure \ref{fig:attn} shows an example of a section of the crossmodal attention matrix on layer 3 of the $V \rightarrow L$ network of MulT (the original matrix has dimension $T_L \times T_V$; the figure shows the attention corresponding to approximately a 6-sec short window of that matrix). We find that crossmodal attention has learned to attend to meaningful signals across the two modalities. For example, stronger attention is given to the intersection of words that tend to suggest emotions (e.g., ``movie'', ``disappointing'') and drastic facial expression changes in the video (start and end of the above vision sequence). This observation advocates one of the aforementioned advantage of MulT over conventional alignment (see Section \ref{subsec:advantage}): crossmodal attention enables MulT to directly capture potentially long-range signals, including those off-diagonals on the attention matrix.






\section{Discussion}

In the paper, we propose Multimodal Transformer (MulT) for analyzing human multimodal language. At the heart of MulT is the crossmodal attention mechanism, which provides a latent crossmodal adaptation that fuses multimodal information by directly attending to low-level features in other modalities. Whereas prior approaches focused primarily on the aligned multimodal streams, MulT serves as a strong baseline capable of capturing long-range contingencies, regardless of the alignment assumption. Empirically, we show that MulT exhibits the best performance when compared to prior methods.

We believe the results of MulT on unaligned human multimodal language sequences suggest many exciting possibilities for its future applications (e.g., Visual Question Answering tasks, where the input signals is a mixture of static and time-evolving signals). We hope the emergence of MulT could encourage further explorations on tasks where alignment used to be considered necessary, but where crossmodal attention might be an equally (if not more) competitive alternative.

\clearpage

\section*{Acknowledgements}

This work was supported in part by DARPA HR00111990016, AFRL FA8750-18-C-0014, NSF IIS1763562 \#1750439 \#1722822, Apple, Google focused award, and Samsung. We would also like to acknowledge NVIDIA’s GPU support.

\bibliography{acl2019}
\bibliographystyle{acl_natbib}

\clearpage

\appendix

\section{Positional Embedding}
\label{app:pe}

A purely attention-based transformer network is \emph{order-invariant}. In other words, permuting the order of an input sequence does not change transformer's behavior or alter its output. One solution to address this weakness is by embedding the positional information into the hidden units~\citep{vaswani2017attention}.

Following \cite{vaswani2017attention}, we encode the positional information of a sequence of length $T$ via the $\sin$ and $\cos$ functions with frequencies dictated by the feature index. In particular, we define the positional embedding (PE) of a sequence $X \in \mathbb{R}^{T \times d}$ (where $T$ is length) as a matrix where:
\begin{align*}
\text{PE}[i, 2j] &= \sin\bigg(\frac{i}{10000^{\frac{2j}{d}}}\bigg) \\
\text{PE}[i, 2j+1] &= \cos\bigg(\frac{i}{10000^{\frac{2j}{d}}}\bigg)
\end{align*}
for $i=1, \dots, T$ and $j=0, \lfloor \frac{d}{2} \rfloor$. Therefore, each feature dimension (i.e., column) of PE are positional values that exhibit a sinusoidal pattern.  Once computed, the positional embedding is added directly to the sequence so that $X + \text{PE}$ encodes the elements' position information at every time step.

\section{Connectionist Temporal Classification}
\label{app:ctc}

Connectionist Temporal Classification (CTC)~\citep{graves2006connectionist} was first proposed for unsupervised Speech to Text alignment. Particularly, CTC is often combined with the output of recurrent neural network, which enables the model to train end-to-end and simultaneously infer speech-text alignment without supervision. For the ease of explanation, suppose the CTC module now are aiming at aligning an audio signal sequence [$a_1, a_2, a_3, a_4, a_5, a_6$] with length $6$ to a textual sequence ``I am really really happy'' with length $5$. In this example, we refer to audio as the source and texts as target signal, noting that the sequence lengths may be different between the source to target; we also see that the output sequence may have repetitive element (i.e., ``really''). The CTC~\cite{graves2006connectionist} module we use comprises two components: alignment predictor and the CTC loss. 

First, the alignment predictor is often chosen as a recurrent networks such as LSTM, which performs on the source sequence then outputs the possibility of being the unique words in the target sequence as well as a empty word (i.e., x). In our example, for each individual audio signal, the alignment predictor provides a vector of length $5$ regarding the probability being aligned to \underline{[x, `I', `am', `really', `happy']}.

Next, the CTC loss considers the negative log-likelihood loss from only the proper alignment for the alignment predictor outputs. The proper alignment, in our example, can be results such as 
\begin{enumerate}[i)]
\item \underline{[x, `I', `am', `really', `really', `happy']}; 
\item \underline{[`I', `am', x, `really', `really', `happy']}; 
\item \underline{[`I', `am', `really', `really', `really', `happy']};
\item \underline{[`I', `I', `am', `really', `really', `happy']}
\end{enumerate}
In the meantime, some examples of the suboptimal/failure cases would be 
\begin{enumerate}[i)]
\item \underline{[x, x, `am', `really', `really', `happy']};
\item \underline{[`I', `am', `I', `really', `really', `happy']};
\item \underline{[`I', `am', x, `really', x, `happy']}
\end{enumerate}
When the CTC loss is minimized, it implies the source signals are properly aligned to target signals. 

To sum up, in the experiments that adopting the CTC module, we train the alignment predictor while minimizing the CTC loss. Then, excluding the probability of blank words, we multiply the probability outputs from the alignment predictor to source signals. The source signal is hence resulting in a pseudo-aligned target singal. In our example, the audio signal is then transforming to a audio signal [$a_1', a_2', a_3', a_4', a_5'$] with sequence length $5$, which is pseudo-aligned to \underline{['I', 'am', 'really', 'really', 'happy']}. 

\newcolumntype{K}[1]{>{\centering\arraybackslash}p{#1}}
\begin{table*}[t!]
\caption{\small Hyperparameters of Multimodal Transformer (MulT) we use for the various tasks. The ``\# of Crossmodal Blocks'' and ``\# of Crossmodal Attention Heads'' are for each transformer.
}
\begin{center}
\fontsize{8}{11}\selectfont
\begin{tabular}{|c||*{3}{K{1.8cm}}|}
\hline
                            & CMU-MOSEI   &  CMU-MOSI  &   IEMOCAP   \\ \hline \hline
Batch Size                  &   16   &    128    &    32   \\
Initial Learning Rate       &   1e-3   &   1e-3   &    2e-3   \\
Optimizer                   &   Adam   &    Adam   &    Adam    \\
Transformers Hidden Unit Size $d$  &   40    &   40    &    40   \\
\# of Crossmodal Blocks $D$     &   4   &    4   &   4    \\
\# of Crossmodal Attention Heads    &   8  &   10    &   10     \\
Temporal Convolution Kernel Size ($L$/$V$/$A$)     &   (1 \text{ or } 3)/3/3    &    (1 \text{ or } 3)/3/3    &   3/3/5   \\
Textual Embedding Dropout   &   0.3   &   0.2   &   0.3   \\
Crossmodal Attention Block Dropout  &   0.1   &    0.2   &    0.25  \\
Output Dropout              &   0.1   &    0.1   &    0.1  \\
Gradient Clip               &   1.0  &   0.8  &   0.8   \\
\# of Epochs                 &  20   &   100   &   30  \\  \hline
\end{tabular}
\end{center}
\vspace{-5mm}
\label{tbl:hyp}
\end{table*}

\section{Hyperparameters}
\label{app:hyper}

Table \ref{tbl:hyp} shows the settings of the various MulTs that we train on human multimodal language tasks. As previously mentioned, the models are contained at roughly the same sizes as in prior works for the purpose of fair comparison. For hyperparameters such as the dropout rate and number of heads in crossmodal attention module, we perform a basic grid search. We decay the learning rate by a factor of 10 when the validation performance plateaus.

\section{Features}
\label{sec:feat}

The features for multimodal datasets are extracted as follows:
\begin{enumerate}
    \item [-] {\bf Language.} We convert video transcripts into pre-trained Glove word embeddings (glove.840B.300d)~\citep{pennington2014glove}. The embedding is a $300$ dimensional vector.
    \item [-] {\bf Vision.} We use Facet~\citep{emotient} to indicate $35$ facial action units, which records facial muscle movement~\citep{ekman1980facial,ekman1992argument} for representing per-frame basic and advanced emotions.   
    \item [-] {\bf Audio.} We use COVAREP~\citep{degottex2014covarep} for extracting low level acoustic features. The feature includes 12 Mel-frequency cepstral coefficients (MFCCs), pitch tracking and voiced/unvoiced segmenting features, glottal source parameters, peak slope parameters and maxima dispersion quotients. Dimension of the feature is $74$.
    
\end{enumerate}

\end{document}